\documentclass{article}

 \usepackage[main, final]{neurips_2026}

\usepackage[utf8]{inputenc} 
\usepackage[T1]{fontenc}    
\usepackage{hyperref}       
\usepackage{url}            
\usepackage{booktabs}       
\usepackage{amsfonts}       
\usepackage{nicefrac}       
\usepackage{microtype}      
\usepackage{xcolor}         
\usepackage{graphicx}
\usepackage{amsmath}
\usepackage{xspace}
\usepackage{enumitem}
\usepackage{float}
\usepackage{placeins} 
\usepackage{wrapfig}
\usepackage{caption}
\newcommand{\OURS}{3DThinkVLA\xspace}

\title{3DThinkVLA: Endowing Vision-Language-Action Models with Latent 3D Priors via 3D-Thinking-Guided Co-training}

\author{%
  Jiaxin Shi\textsuperscript{1,6},
  Xidong Zhang\textsuperscript{2,6,7},
  Fucai Zhu\textsuperscript{6},
  Zhe Li\textsuperscript{3},
  Siyu Zhu\textsuperscript{4},
  Weihao Yuan\textsuperscript{5,6,*}
  \\[0.5em]
  \textsuperscript{1}Shanghai Jiao Tong University,
  \textsuperscript{2}Harbin Institute of Technology,
  \\
  \textsuperscript{3}Nanyang Technological University,
  \textsuperscript{4}Fudan University,
  \textsuperscript{5}Nanjing University,
  \\
  \textsuperscript{6}Daimon Robotics,
  \textsuperscript{7}Great Bay University
  \\[0.3em]
  \textsuperscript{*}Corresponding author
}

\makeatletter
\renewcommand{\@notice}{}
\makeatother

\begin{document}

\maketitle

\begin{abstract}
  We propose a 3D-thinking-guided co-training framework that enables vision-language-action (VLA) models to perform 3D spatial reasoning implicitly during action prediction. Our core insight is that 3D geometry perception and 3D spatial reasoning are distinct capabilities that can be disentangled and injected at different feature hierarchies. During training, three tightly coupled components work in concert primarily within the latent space: (1) To gain geometric priors, a \textbf{latent 3D geometry perception module} aligns intermediate visual features with a 3D foundation model, acquiring low-level geometric cues without requiring architectural modifications to the VLM backbone. (2) Complementing this perception, an \textbf{online 3D reasoning distillation module} mitigates the prompt-induced reasoning gap via a shared reasoning anchor token. During 3D VLM co-training, this anchor is emitted as the first output token to robustly encode spatial priors. During VLA training, it serves as an input token inserted between the task and action instructions, transferring high-level spatial thinking from explicit teacher reasoning prompts to student action prompts without requiring chain-of-thought text generation. (3) These disentangled geometric and reasoning features are then united by a \textbf{spatially augmented action integration}, which jointly injects them into the action-query tokens as hierarchical spatial conditions to prevent action shortcuts. At deployment, our method retains only its lightweight adapters to perform implicit 3D reasoning, discarding the 3D foundation model and the teacher branch used for supervision. Consequently, it operates purely on 2D images without requiring 3D sensors, external foundation models, or explicit text generation while effectively preventing catastrophic forgetting of the pretrained VLM, achieving state-of-the-art performance on LIBERO, LIBERO-PLUS, SimplerEnv, and challenging real-world manipulation tasks.
  The code of our method will be made public.

\end{abstract}

\begin{figure*}[]
    \centering
    \includegraphics[width=1\linewidth]{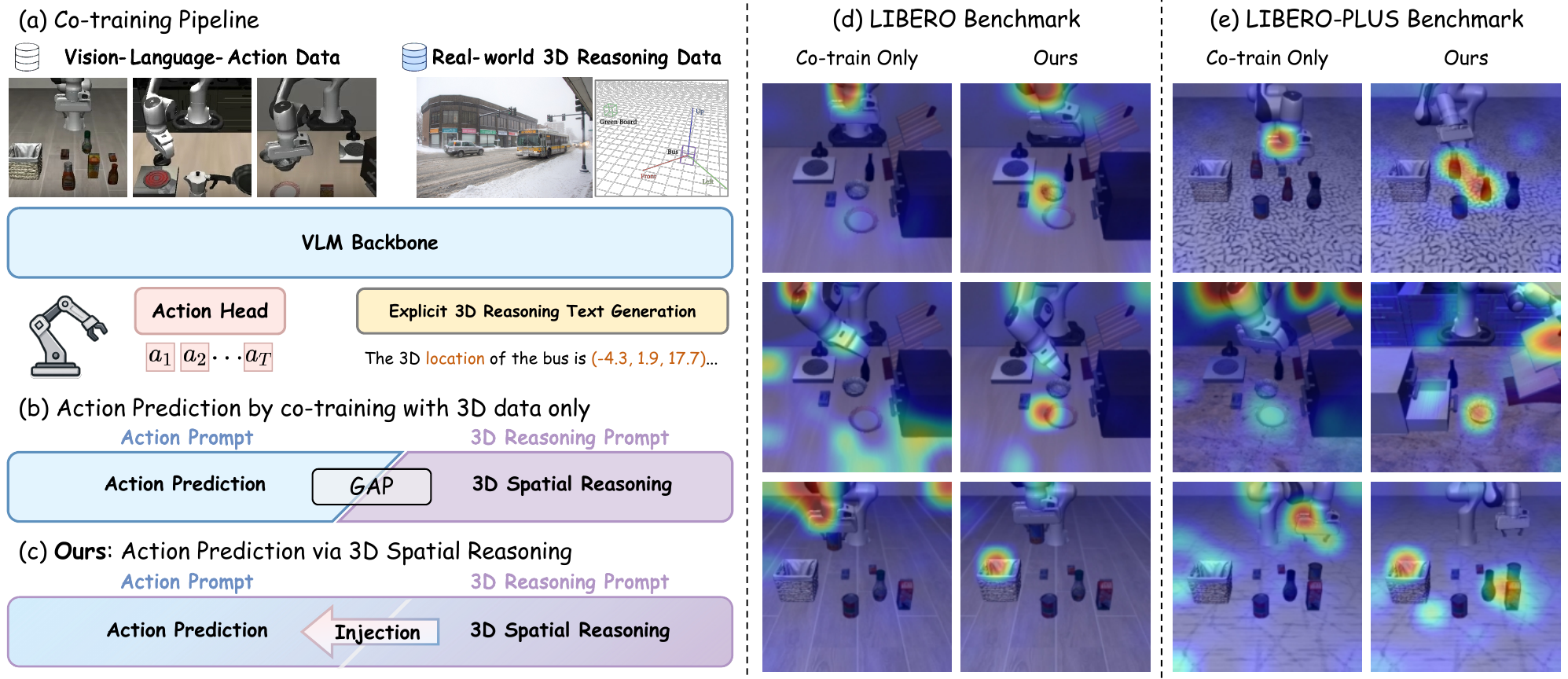}
    \caption{Overview. (a) Our framework co-trains the VLM backbone on VLA data and real-world 3D reasoning data to enhance embodied spatial intelligence. (b) We identify a prompt-induced reasoning gap in vanilla co-training, where the model tends to learn action shortcuts and deactivates its spatial perception when triggered by action prompts. (c) Our method bridges this gap by utilizing a reasoning adapter and online latent distillation to inject 3D spatial thinking into action prediction without requiring explicit text generation. (d)-(e) Qualitative results on LIBERO and LIBERO-PLUS benchmarks demonstrate that while the co-train-only model exhibits scattered attention or focuses on task-irrelevant areas like the robot arm, our method focuses precisely on task-relevant objects and their 3D spatial structures, leading to more robust manipulation.}
    \label{fig:intro}
    \vspace*{-0.5cm}
\end{figure*}

\section{Introduction}

Vision-language-action (VLA) models aim to enable embodied agents to predict actions and interact within physical 3D environments~\cite{zitkovich2023rt,brohan2022rt,zhang2026vlm4vla}. However, most existing VLAs predominantly rely on 2D images as visual inputs, creating a critical gap between 2D semantics and 3D spatial reasoning~\cite{3ddifusionpolicy,feng2025spatial,sun2025geovla}.

To bridge this gap, one line of work injects explicit 3D information (e.g., point clouds, depth maps) into the action head~\cite{li2026pointvla,sun2025geovla,zhang2025spatial} or VLM backbone~\cite{qu2025spatialvla,qian2025geopredict}, or uses it as auxiliary prediction targets~\cite{3dvla,yuan2025depthvla,song2026robotic,li2025qdepth}. While effective at enhancing 3D perception, these methods typically depend on external 3D sensors or require non-trivial modifications to the VLM backbone. Another line of work implicitly integrates 3D geometry-aware information by aligning VLM visual features with 3D foundation models~\cite{li2025spatial,zhong2026dualcot}. However, such direct feature alignment risks disrupting the VLM's original vision-language alignment. Moreover, both families of methods focus on low-level geometry injection. They fall short of providing the high-level 3D spatial reasoning, like inferring object positions and relative orientations, that robotic manipulation demands.

To endow models with spatial reasoning, some works incorporate 3D data during pre-training~\cite{zhang2026vlm4vla,3dvla}. However, the domain gap between 3D pre-training data and downstream VLA tasks often leads to catastrophic forgetting during fine-tuning. A natural remedy is co-training on both action data and 3D data, which we find effectively mitigates forgetting and improves downstream performance. Yet this strategy reveals a more insidious problem: a prompt-induced reasoning gap. Specifically, while standard 3D visual question answering (VQA) prompts successfully activate the model's spatial reasoning, simple action-prediction prompts cause the model to bypass these learned spatial priors and revert to an action shortcut, as shown in Figure~\ref{fig:intro}.

To overcome these limitations, we propose \textbf{\OURS}, a 3D-thinking-guided co-training framework that enables VLAs to perform 3D spatial reasoning implicitly during action prediction. Our core insight is that 3D geometry perception and 3D spatial reasoning are distinct capabilities that can be disentangled and injected at different feature hierarchies of the model. Concretely, we co-train the model on VLA data and 3D reasoning data annotated on real-world images. During action training, three tightly coupled components work in concert: (1) \textbf{Latent 3D Geometry Perception Module}. To equip the VLA with geometric priors without modifying the VLM backbone, we extract intermediate visual features from the vision encoder and pass them through a lightweight geometry adapter. This adapter aligns the visual features with a 3D foundation model in latent space, capturing low-level geometry cues. (2) \textbf{Online 3D Reasoning Distillation Module}. To close the prompt-induced reasoning gap, we introduce a teacher-student distillation scheme operating entirely in latent space. We design a special reasoning token and insert it after the task instruction in both teacher and student branches; in the co-training VLM stream, we also require this reasoning token to be emitted as the first output token. The teacher branch processes 3D spatial reasoning prompts to produce 3D reasoning token representations. The student branch, conditioned on standard action prompts, maps its reasoning tokens into a reasoning latent space via a lightweight reasoning adapter and is trained to match the teacher's representations. This transfers high-level 3D reasoning priors into the action-prediction pathway without requiring explicit chain-of-thought (CoT) text generation. (3) \textbf{Spatially Augmented Action Integration}. To further alleviate action shortcuts and fully exploit the learned spatial knowledge, we inject the geometry-aware features from the perception module and the reasoning-aligned tokens from the distillation module jointly into the action head as hierarchical spatial conditions, providing the action head with complementary low-level geometric and high-level reasoning knowledge for robust prediction.

During inference, our model requires no explicit 3D sensor inputs, no external 3D foundation models, and no CoT text generation. The framework is architecture-agnostic and integrates into existing VLA pipelines with minimal modifications while fully preserving the pretrained VLM's semantic alignment.
Extensive experiments demonstrate our method achieves state-of-the-art performance on LIBERO, LIBERO-PLUS, SimplerEnv, and challenging real-world manipulation tasks.

The main contributions of this work are summarized as follows:
\begin{itemize}[leftmargin=*]
    \item  We identify and characterize the prompt-induced reasoning gap in VLA co-training, revealing that standard action prompts can deactivate learned spatial priors and lead to sub-optimal action shortcuts.
    \item We propose \OURS, a 3D-thinking-guided co-training framework that disentangles 3D geometry perception from spatial reasoning via three tightly coupled components: a latent geometry perception module, an online reasoning distillation module, and a spatially augmented action integration module.
    \item We introduce an online 3D reasoning distillation mechanism using a shared reasoning anchor token, enabling the transfer of high-level spatial thinking from reasoning prompts to action prompts entirely in the latent space, bypassing the need for explicit text generation.
    \item Extensive experiments demonstrate that our method achieves state-of-the-art performance on LIBERO, LIBERO-PLUS, SimplerEnv, and real-robot benchmarks while enabling 3D-input-free inference.
\end{itemize}
\section{Related Works}

\textbf{3D vision language action model.}
Recent 3D-aware policies address the spatial grounding limitations of 2D VLAs~\cite{brohan2022rt,zitkovich2023rt,zhang2026vlm4vla}. Explicit methods inject 3D representations (e.g., point clouds) into VLM backbones~\cite{qu2025spatialvla,3dvla,nikolov2026spear1scalingrobotdemonstrations} or action heads~\cite{li2026pointvla,sun2025geovla,yang2025fp33dfoundationpolicy}, though often constrained by sensor requirements or semantic misalignment. To achieve 3D-free inference, implicit approaches leverage 2D-3D projection~\cite{li2025bridgevlainputoutputalignmentefficient}, feature lifting~\cite{jia2024lift3dfoundationpolicylifting}, or distillation from 3D foundation models~\cite{abouzeid2025geoaware, li2025spatial,zhang2026spatialactionsgroundingvisionlanguageaction, wang2025vggt}. Advanced works also incorporate predictive 3D dynamics via depth prediction~\cite{li2025qdepth}, spatiotemporal calibration~\cite{4dvla,niu2025pretrainingautoregressiveroboticmodels}, or scene forecasting~\cite{qian2025geopredict,3dvla}. Unlike these, our method disentangles geometry perception from spatial reasoning, injecting both as hierarchical conditions. By bridging the prompt-induced reasoning gap via latent distillation, we enable 3D-aware prediction from 2D images without specialized sensors, external models, or explicit CoT.

\textbf{Self-distillation.} Knowledge distillation has recently emerged as an important technique for optimizing VLA models, primarily focusing on two directions: computational efficiency and policy refinement. To reduce inference latency, ActDistill~\cite{ye2026actdistillgeneralactionguidedselfderived} and AC$^2$-VLA~\cite{yu2026ac2vlaactioncontextawareadaptivecomputation} employ action-guided self-distillation to train adaptive computation mechanisms, such as dynamic routing and token pruning. To enhance closed-loop robustness and bridge the gap between offline SFT and online RL, VLA-OPD~\cite{zhong2026vlaopdbridgingofflinesft} and PLD~\cite{xiao2025selfimprovingvisionlanguageactionmodelsdata} distill expert knowledge or residual RL behaviors onto self-generated trajectories. Similarly, EvoDriveVLA~\cite{cao2026evodrivevlaevolvingautonomousdriving} and VITA-VLA~\cite{dong2025vitavlaefficientlyteachingvisionlanguage} utilize distillation to transfer privileged planning information or pretrained action-expert capabilities into VLM backbones. In contrast to these methods, \OURS leverages distillation to mitigate the prompt-induced reasoning gap. Rather than distilling explicit actions, our method performs online latent self-distillation, implicitly injecting high-level spatial reasoning ability into the action prediction pathway.
\begin{figure*}[]
    \centering
    \includegraphics[width=1\linewidth]{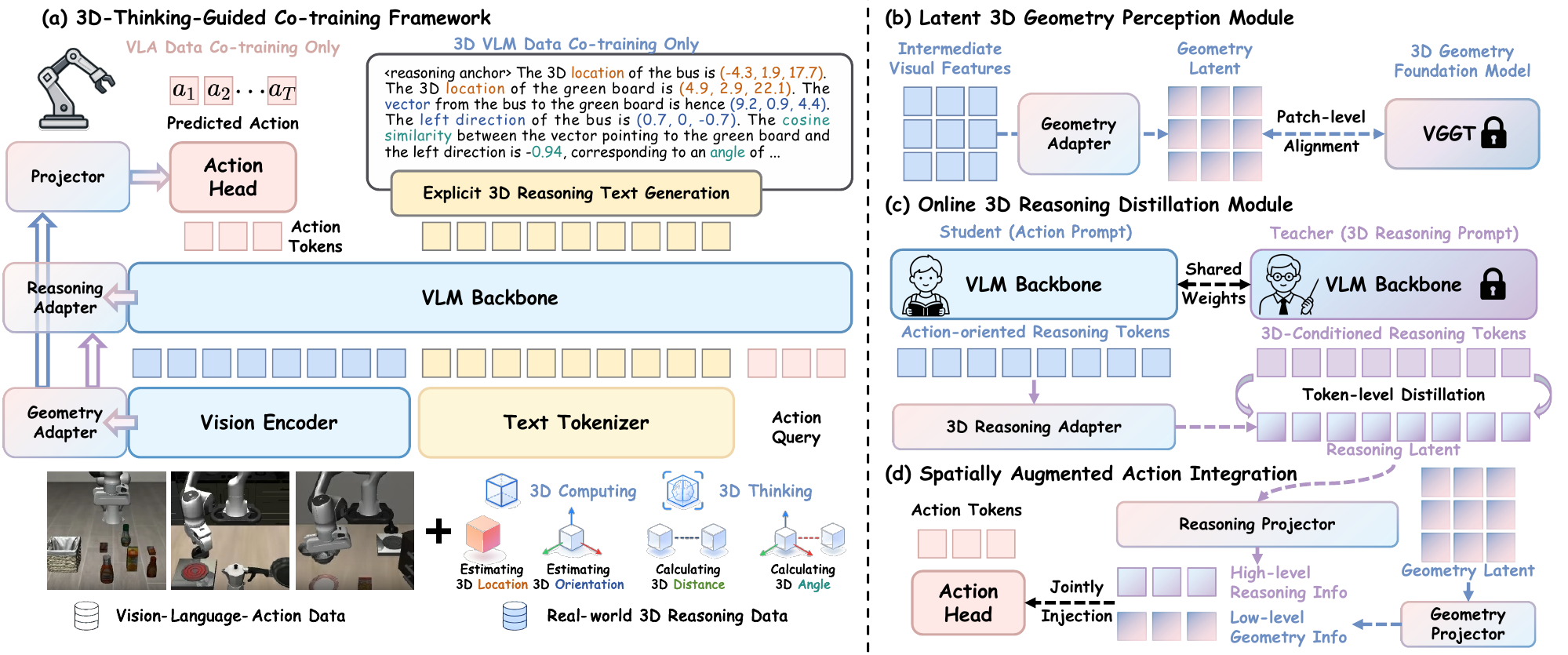}
    \vspace{-6mm}
    \caption{
    Our 3D-thinking-guided framework disentangles geometric perception and spatial reasoning to enhance VLA models. (a) The VLM backbone is co-trained on VLA and 3D data, internalizing spatial intelligence through explicit 3D reasoning tasks. (b) A Geometry Adapter aligns visual features with a 3D foundation model (VGGT) via patch-level latent alignment to capture low-level geometric priors. (c) Online distillation transfers high-level reasoning from teacher prompts to student action prompts via a shared anchor token, bridging the prompt-induced reasoning gap. (d) Disentangled reasoning and geometry features are jointly injected into the action head as hierarchical conditions to guide prediction and prevent action shortcuts. At inference, auxiliary modules are discarded, enabling efficient implicit 3D reasoning from 2D images via lightweight adapters.
    }
    \label{fig:overall}
    \vspace{-4mm}
\end{figure*}

\section{Method}
We present \OURS, a 3D-thinking-guided VLA co-training framework as shown in Figure~\ref{fig:overall}. The key idea is to enrich action prediction with two complementary latent cues during training: (i) low-level geometry-aware cues distilled from the 3D geometry foundation model and (ii) high-level reasoning-aware cues transferred from a teacher reasoning pathway and online self-distillation strategy. 
Importantly, these auxiliary supervision sources (\textit{i.e.}, the 3D foundation model and the teacher reasoning pathway) are utilized only during training. At deployment, only the lightweight geometry and reasoning adapters are retained. This ensures the inference-time interface remains as efficient as a standard VLA, requiring no extra 3D sensor or auxiliary models.

Our method contains three tightly coupled components: 
(1) a Latent 3D Geometry Perception Module that equips the VLA with geometric priors by aligning intermediate visual features with a 3D foundation model via a lightweight geometry adapter;
(2) an Online 3D Reasoning Distillation Module that mitigates the prompt-induced reasoning gap by transferring spatial reasoning from explicit teacher prompts to student action prompts via the reasoning adapter and a shared reasoning anchor; and
(3) a Spatially Augmented Action Integration Module that unites these disentangled geometric and reasoning features, injecting them into the action head as hierarchical spatial conditions.

\subsection{Preliminary}
\label{sec:preliminary}
\textbf{Problem definition.} The VLA paradigm aims to learn a policy $\pi$ that maps the current multi-view observations $I_t$, the language instruction $L$, and action queries $\tau_A$ to the robot's action chunk with $H$ horizons $A_t=[a_t, a_{t+1}, \cdots, a_{t+H-1}]$:
\begin{equation}
    \hat{A}_t=\pi_\Theta(I_t,L, \tau_A).
\end{equation}
Each action $a_t \in \mathbb{R}^7$ is a 7-DoF end-effector command $a_t=\{\delta x, \delta \theta, g\}$, where $\delta x \in \mathbb{R}^3$ and $\delta \theta \in \mathbb{R}^3$ denote translational and rotational offsets, and $g \in \mathbb{R}$ is the gripper's open/close state. Instruction $L$ is composed of the task instruction $L_{\text{task}}$ and the action-prediction instruction $L_{\text{action}}$. Although VLA models are initialized from a pretrained VLM, they still lack strong 3D spatial reasoning ability and tend to forget pretrained VLM knowledge during action fine-tuning. Thus, \OURS aims to inject high-level 3D reasoning and low-level 3D geometry priors into the VLA model to enable 3D spatial reasoning and, at the same time, keep the original VLM knowledge. 

\textbf{Backbone and action head.}
We adopt Qwen3-VL-2B as the backbone of our VLA model and employ an OFT-style action head for action prediction. Language, image, and action-query tokens are first projected into a shared latent space and processed by the transformer; the action head then produces $\hat{A}_t$ by aggregating the hidden states corresponding to the action-query tokens.

\subsection{Latent 3D Geometry Perception}
\label{sec:latent_3d_geometry_perception}

\textbf{Geometry adapter.}
Considering the modality gap between 3D geometric information and the pretrained vision-language alignment of the VLM backbone, we do not directly align geometry features with visual tokens. Instead, we design a geometry adapter $\mathcal{G}$, consisting of an MLP and LayerNorm, which serves as a bridge between geometry information and visual semantic features. During training, we extract and leverage intermediate visual features specifically from the 18th layer of the vision encoder, thereby avoiding repeated feature extraction. Given the patch-level intermediate visual feature $\mathcal{F}_v \in \mathbb{R}^{B \times C \times H_v \times W_v}$, where $C$ is the channel dimension, $H_v$ and $W_v$ are the height and width of the feature map, we first apply a lightweight MLP to project the visual feature into the geometry latent space and align the projected features with the 3D foundation model, i.e., VGGT~\cite{wang2025vggt}, which outputs the 3D geometric features $\mathcal{F}^{3D} \in \mathbb{R}^{B \times C_f \times H_f \times W_f}$, where $C_f$ is the channel dimension, $H_f$ and $W_f$ are the height and width of the feature map. 

\textbf{Alignment.}
To perform patch-level alignment, we use bilinear interpolation to resize the 3D geometric features to the same resolution as the intermediate visual features. 
\begin{equation}
\mathcal{L}_{\text{geo}} = 1 - \mathcal{S}(\mathcal{F}^{3D}, \mathcal{F}^{\text{Geo}}),
\end{equation}
where $\mathcal{S}$ denotes cosine similarity, $\mathcal{F}^{\text{Geo}}=\mathcal{G}\left(\mathcal{F}_v \right)$ is the geometry adapter output.

\subsection{Online 3D Reasoning Distillation}
\label{sec:online_3d_reasoning_distillation}
Due to the domain gap between the action space and the vision-language space, a fine-tuned VLA model can catastrophically forget the knowledge of its pretrained VLM backbone. In addition, during action prediction, attention may become diffuse and drift toward task-irrelevant regions. As a result, the model cannot effectively leverage the strong generalization ability of the VLM and suffers from reduced universality. To address this issue, we propose an online teacher-student distillation scheme that preserves spatial reasoning patterns and spatial affordance awareness during action prediction.

\textbf{Teacher and student reasoning anchors.}
We use the teacher branch to supervise the student branch in spatial reasoning. To establish a consistent bottleneck for distillation across varying prompt formats, we introduce a shared reasoning-anchor token $\tau_R$ to align their reasoning and action-prediction behaviors. Specifically, we insert this token after the task instruction. Under causal attention, this position enables the reasoning anchor to absorb both visual observations and task semantics while remaining minimally affected by downstream action decoding. The action-query tokens are placed after the action-prediction instruction so that they can attend to, and benefit from, the reasoning information carried by the anchor. The action prediction can be reformulated as
\begin{equation}
\hat{A}_t=\pi_\Theta(I_t,L_{\text{task}}, \tau_R, L_{\text{action}}, \tau_A).
\end{equation}
Although the student branch already includes a reasoning anchor, action-oriented prompting $L_{\text{action}}$ can still yield weaker reasoning activations than explicit reasoning prompts. During training, we therefore construct a teacher instruction $L^{\text{teacher}}$ to elicit spatial reasoning from the VLM, and we gather the hidden states of the teacher's reasoning anchor:
\begin{equation}
H_{\text{teacher}}^R = \text{sg}\!\left(f_\theta(I_t, L_{\text{task}}, L_{\text{teacher}}, \tau_{R})\right),
\end{equation}
where $f_\theta$ is a function that gathers the last hidden state of the shared reasoning anchor $\tau_R$. $H_\text{teacher}^R$ is the 3D-conditioned reasoning tokens elicited by the 3D reasoning prompt. $\text{sg}(\cdot)$ represents stop gradients. Also, we can obtain the student reasoning anchor information from the student branch:
\begin{equation}
\hat{H}_{\text{student}}^R=f_\theta(I_t,L_{\text{task}}, \tau_R, L_{\text{action}}, \tau_A),
\end{equation}
where $\hat{H}_{\text{student}}^R$ represents the action-oriented reasoning tokens. Notably, for the reasoning anchor, the only difference between the teacher and student branches is the instruction language; they share the same visual observations. In this process, the teacher and student branches share parameters, and we do not require slow next-token generation for alignment or any additional model to provide distillation signals. To maintain the teacher branch's spatial reasoning capability and accelerate training, we stop gradients through the teacher branch.

\textbf{Distillation loss.}
If we directly align the teacher and student reasoning anchors, it may cause a gradient explosion and degrade task-instruction following. We thus design a reasoning adapter $\mathcal{R}$ to project the hidden state of the student's reasoning anchor into a latent reasoning space. We then perform token-level distillation to inject the teacher's spatial reasoning knowledge into this latent reasoning space. This distillation process encourages the action-oriented tokens to inherit spatial reasoning patterns from the 3D-oriented teacher tokens, even without explicit reasoning instructions in the student prompt,
\begin{equation}
    \mathcal{L}_{\text{reasoning}}= 1 - \mathcal{S}(H_\text{teacher}^R, \mathcal{R}(\hat{H}_{\text{student}}^R)),
\end{equation}
where $\mathcal{R}$ uses a lightweight MLP with LayerNorm for simplicity.

Since the teacher shares parameters with the student, the token-feature distributions remain similar, which improves distillation stability during training and better exploits the pretrained model's intrinsic capability. Meanwhile, this design may cause feature collapse of the reasoning token, making it encode less useful information; we address this issue in Section~\ref{sec:training_and_inference} via reasoning-grounded co-training. Notably, at inference time, we no longer need the teacher branch.

\subsection{Spatially Augmented Action Integration}
\label{sec:spatially_augmented_action_integration}
\textbf{Query-level spatial injection.}
In Sections~\ref{sec:latent_3d_geometry_perception} and~\ref{sec:online_3d_reasoning_distillation}, we use a geometry adapter and a reasoning adapter to learn geometry priors and reasoning knowledge in their corresponding latent spaces. To better connect knowledge from these latent spaces with action prediction, we design Spatially Augmented Action Integration, which injects both signals into the action head to guide prediction together with the action-query tokens.
We first use two additional lightweight MLP layers to project features from the geometry latent space and the reasoning latent space into the action latent space, producing $H_\text{geo}^A$ and $H_\text{reasoning}^A$. Then, we inject these features into the action-query token $\tau_A$ through element-wise addition:
\begin{equation}
    \hat{A}_{t} = \text{Action Head} (H_A + H_\text{geo}^A + H_\text{reasoning}^A),
\end{equation}
where $H_A$ denotes the last hidden state of $\tau_A$.

To balance the influences of $H_A$, $H_\text{geo}^A$, and $H_\text{reasoning}^A$ on action prediction and to reduce overfitting, we randomly drop $H_\text{geo}^A$ and $H_\text{reasoning}^A$ on a subset of training samples. We empirically find that a simple additive fusion performs best, as discussed in Appendix~\ref{sec:reasoing_fusion}.

\subsection{Training Objective and Inference Path}
\label{sec:training_and_inference}
We train the model with a co-training strategy, jointly using both VLA data and VLM data. Specifically, to equip the model with 3D reasoning ability, we include a 3D-reasoning VLM dataset in co-training, which also strengthens the teacher branch's guidance of the student branch for 3D spatial reasoning. In each training step, we perform two forward passes, one with VLM data and one with VLA data, and accumulate gradients from both before a single backward update.

\textbf{VLA-step objective.}
Within the VLA stage, the framework computes action loss, geometry alignment loss, and distillation loss. The objective function is
\begin{equation}
\mathcal{L}_{\text{vla}} = \mathcal{L}_{\text{action}} + \lambda_a\,\mathcal{L}_{\text{geo}} + \lambda_d\,\mathcal{L}_{\text{reasoning}},
\label{eq:vla_step}
\end{equation}
where $\lambda_a$ and $\lambda_d$ are auxiliary loss weights. For action prediction, we calculate the L1 distance between the predicted action chunk and the ground truth action chunk.

\textbf{Co-training with the 3D VLM stream.}
The 3D VLM training data consists of real-world images paired with dialogue and question-answer (QA) text. These data require explicit 3D spatial reasoning (e.g., 3D bounding boxes) as well as relational 3D reasoning (e.g., distance and orientation). For more details about 3D co-training data see Appendix~
\ref{sec:cotrain_data}. To better leverage general 3D reasoning learned through co-training, we further adapt the VLM training format by requiring the model to emit the reasoning-anchor token as the first output token. Due to the autoregressive next-token prediction mechanism, subsequent 3D-reasoning text is then generated conditioned on this reasoning anchor, strengthening its representation of 3D spatial reasoning and avoiding feature collapse of the reasoning token.
Training, therefore, jointly accumulates action-centric supervision and language-reasoning supervision, consistent with the dual-dataloader co-training loop. In addition to the VLA-stage objective, each iteration performs a second forward pass on a VLM batch using the shared Qwen3-VL-2B backbone:
\begin{equation}
\mathcal{L}_{\text{vlm}}= \lambda_{\text{3D}} \, \mathcal{L}_{\text{CE}},
\end{equation}
where $\lambda_{\text{3D}}$ is the loss weight for 3D VLM data and $\mathcal{L}_{\text{CE}}$ denotes the standard cross-entropy loss for autoregressive language modeling. The total loss is  $\mathcal{L}_{\text{total}} = \mathcal{L}_{\text{vla}} + \mathcal{L}_{\text{vlm}}$. We also analyze the gradient via co-training in
Appendix~\ref{sec:gradient}.

\begin{table*}[]
\centering
\small
\caption{\textbf{LIBERO Benchmark Results.} Success rates (\%) across 4 evaluation suites are presented. \textbf{Bold} indicates the \textbf{best}-performing model, and \underline{underline} indicates the \textbf{runner-up} model.}
\vspace{-0mm}
\resizebox{\textwidth}{!}{%
\begin{tabular}{l | c  c  c  c | c}
\toprule
\textbf{Method} & 
\multicolumn{1}{c}{\textbf{LIBERO-Spatial}} & 
\multicolumn{1}{c}{\textbf{LIBERO-Object}} & 
\multicolumn{1}{c}{\textbf{LIBERO-Goal}} & 
\multicolumn{1}{c|}{\textbf{LIBERO-Long}} & 
\multicolumn{1}{c}{\textbf{Average}} \\
\midrule
TraceVLA~\citep{tracevla} & 84.6 & 85.2 & 75.1 & 54.1 & 74.8 \\
Octo~\citep{octo} & 78.9 & 85.7 & 84.6 & 51.1 & 75.1 \\
OpenVLA~\citep{openvla} & 84.7 & 88.4 & 79.2 & 53.7 & 76.5 \\
CoT-VLA~\citep{cot-vla} & 87.5 & 91.6 & 87.6 & 69.0 & 83.9 \\
$\pi_0$-FAST~\citep{pi0-fast} & 96.4 & 96.8 & 88.6 & 60.2 & 85.5 \\
$\pi_0$~\citep{pi0} & 96.8 & 98.8 & 95.8 & 85.2 & 94.2\\
OpenVLA-OFT~\citep{openvla-oft} & 97.6 & 98.4 & 97.9 & 94.5 & 97.1 \\
SpatialVLA~\citep{qu2025spatialvla} & 88.2 & 89.9 & 78.6 & 55.5 & 78.1 \\
GeoVLA~\citep{geovla} & 98.4 & 99.0 & 96.6 & \underline{96.6} & 97.7 \\
3D-CAVLA~\citep{bhat2026} & 98.2 & \underline{99.8} & \underline{98.2} & 96.1 & 98.1 \\
SpatialForcing~\cite{li2025spatial} & \underline{99.4} & 99.6 & \textbf{98.8} & 96.0 & \underline{98.5} \\
VITA~\cite{vita} & 95.9 & 98.9 & 95.1 & \textbf{96.8} & 96.7\\
\midrule
\textbf{Ours} & \textbf{100.0} & \textbf{100.0} & \textbf{98.8} & 95.8 & \textbf{98.7} \\
\bottomrule
\end{tabular}%
}
\label{tab:libero}
\vspace{-3mm}
\end{table*}

\begin{table*}[]
\centering
\small
\caption{\textbf{LIBERO-Plus Benchmark Results.} Success rates (\%) on 7 perturbation dimensions are presented. \textbf{Bold} indicates the \textbf{best}-performing model, and \underline{underline} indicates the \textbf{runner-up} model.}
\vspace{-0mm}
\resizebox{\textwidth}{!}{%
\begin{tabular}{l | c c c c c c c | c}
\toprule
\textbf{Method} &
\multicolumn{1}{c}{\textbf{Camera}} &
\multicolumn{1}{c}{\textbf{Robot}} &
\multicolumn{1}{c}{\textbf{Language}} &
\multicolumn{1}{c}{\textbf{Light}} &
\multicolumn{1}{c}{\textbf{Background}} &
\multicolumn{1}{c}{\textbf{Noise}} &
\multicolumn{1}{c|}{\textbf{Layout}} &
\multicolumn{1}{c}{\textbf{Avg}} \\
\midrule
OpenVLA~\cite{openvla} & 0.8 & 3.5 & 23.0 & 8.1 & 34.8 & 15.2 & 28.5 & 15.6 \\
OpenVLA-OFT~\cite{openvla-oft} & 56.4 & 31.9 & 79.5 & 88.7 & 93.3 & 75.8 & 74.2 & 69.9 \\
NORA~\cite{nora} & 2.2 & 37.0 & 65.1 & 45.7 & 58.6 & 12.8 & 62.1 & 39.0 \\
WorldVLA~\cite{worldvla} & 0.1 & 27.9 & 41.6 & 43.7 & 17.1 & 10.9 & 38.0 & 25.0 \\
$\pi_0$~\cite{pi0} & 13.8 & 6.0 & 58.8 & 85.0 & 81.4 & 79.0 & 68.9 & 53.6 \\
ABot-M0~\cite{abotm0} & \underline{60.4} & \textbf{67.9} & \underline{86.4} & 96.2 & 91.6 & \textbf{86.4} & \textbf{82.6} & \underline{80.5} \\
Qwen3-VL-OFT~\cite{ye2026starvla} & 47.0 & \underline{60.1} & \textbf{87.0} & \underline{96.3} & \textbf{95.3} & 73.1 & 79.2 & 75.0 \\
\midrule
\textbf{Ours} & \textbf{73.8} & 64.5 & 78.0 & \textbf{98.4} & \underline{94.8} & \underline{84.7} & \underline{81.5} & \textbf{81.0} \\
\bottomrule
\end{tabular}%
}
\label{tab:libero_plus}
\vspace{-3mm}
\end{table*}

\section{Experiments} \label{sec:experiments}

\vspace{-2mm}
\subsection{Experimental Setup} 
\vspace{-2mm}
\label{sec:exp_setup}

We adopt the StarVLA framework~\cite{ye2026starvla}, utilizing Qwen3-VL 2B~\cite{bai2025qwen3vltechnicalreport} as the base model and using OFT-style action head for action prediction. All models are trained on 8 NVIDIA A100 (80GB) GPUs, and the fine-tuned models are deployed on a single NVIDIA A100 (80GB) GPU for evaluation. We evaluate our method on three widely used simulation benchmarks: LIBERO~\cite{liu2023libero}, LIBERO-PLUS~\cite{fei2025libero}, and SimplerEnv~\cite{li24simpler}. 
For further details regarding the implementation and simulation benchmarks, please refer to Appendices~\ref{app:implementation} and~\ref{app:simulation}.

\begin{figure*}[t]
    \centering
    \includegraphics[width=\linewidth]{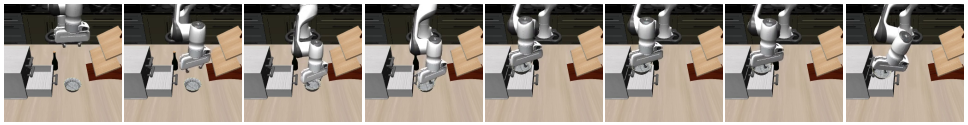}\par\nointerlineskip\vspace{0.5mm}
    \centerline{\scriptsize Task 1 Qwen3-VL-OFT}\par\vspace{1mm}
    \includegraphics[width=\linewidth]{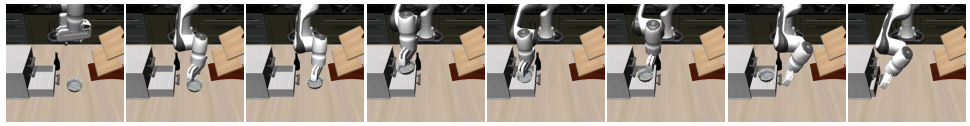}\par\nointerlineskip\vspace{0.5mm}
    \centerline{\scriptsize Task 1 Ours}\par\vspace{1mm}
    \includegraphics[width=\linewidth]{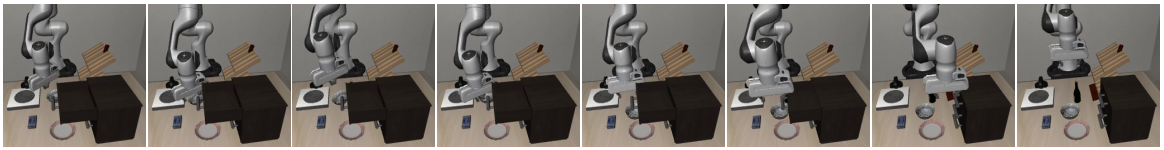}\par\nointerlineskip\vspace{0.5mm}
    \centerline{\scriptsize Task 2 Qwen3-VL-OFT}\par\vspace{1mm}
    \includegraphics[width=\linewidth]{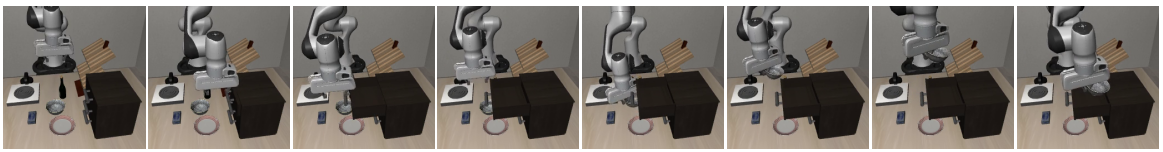}\par\nointerlineskip\vspace{0.5mm}
    \centerline{\scriptsize Task 2 Ours}
    \caption{Qualitative Results on LIBERO-Plus. {Task 1}: ``Put the black bowl in the bottom drawer of the cabinet and close it.'' {Task 2}: ``Put the black bowl in the top drawer of the cabinet and close it.''}
    \vspace{-3mm}
    \label{fig:qualitative}
\end{figure*}

\vspace{-2mm}
\subsection{Evaluation}
\vspace{-2mm}

\textbf{LIBERO.}
Following previous methods~\cite{abotm0,ye2026starvla}, we train a single model across all task suites, and report the results in Table~\ref{tab:libero}. From the results, our method outperforms previous methods in three suites and achieves the highest average success rate.
Notably, our method attains 100\% success rates in two suites, suggesting its capability extends beyond the inherent challenges of these two suites.

\textbf{LIBERO-Plus.}
After being trained on LIBERO, we directly transfer the model to the LIBERO-Plus benchmark for zero-shot evaluation.
As shown in Table~\ref{tab:libero_plus}, our method achieves the highest average success rate compared to all previous methods.
The average success rate of 81.0\% underscores the strong generalization and robustness of our method to various disturbances.
In particular, compared to baselines lacking explicit spatial understanding, our method exhibits a clear advantage on trajectories involving height variation.
As illustrated in Fig.~\ref{fig:qualitative}, Qwen3-VL-OFT frequently misestimates the height of objects to be interacted with along the trajectory, leading to failures such as collisions with surrounding objects or incorrect judgments of target positions.
In contrast, our method consistently produces accurate height-aware predictions, demonstrating that 3D spatial reasoning is a critical capability for handling such challenging perturbations.

\begin{table*}[]
\centering
\small
\caption{\textbf{SimplerEnv WidowX Benchmark Results.} Success rates (\%) across 4 tasks are presented. \textbf{Bold} indicates the \textbf{best}-performing model while \underline{underline} indicates the \textbf{runner-up} model.}
\vspace{-2mm}
\resizebox{\textwidth}{!}{%
\begin{tabular}{l | c c c c | c}
\toprule
\textbf{Method} &
\multicolumn{1}{c}{\textbf{Put Carrot on Plate}} &
\multicolumn{1}{c}{\textbf{Put Eggplant in Basket}} &
\multicolumn{1}{c}{\textbf{Put Spoon on Towel}} &
\multicolumn{1}{c|}{\textbf{Stack Block}} &
\multicolumn{1}{c}{\textbf{Avg}} \\
\midrule
Octo~\cite{octo} & 8.3 & 43.1 & 12.5 & 0.0 & 16.0 \\
OpenVLA~\cite{openvla} & 0.0 & 4.1 & 0.0 & 0.0 & 1.0 \\
RoboVLM~\cite{li2023generalist} & 20.8 & 79.2 & 45.8 & 4.2 & 37.5 \\
SpatialVLA~\cite{qu2025spatialvla} & 25.0 & \textbf{100.0} & 16.7 & 29.2 & 42.7 \\
Open $\pi_0$~\cite{pi0} & 61.3 & 89.6 & 73.7 & 15.8 & 60.0 \\
QDepth-VLA~\cite{li2025qdepth} & 57.5 & 95.0 & 82.0 & \underline{39.6} & 68.5 \\
FALCON~\cite{zhang2026spatialactionsgroundingvisionlanguageaction} & 41.7 & \textbf{100.0} & 62.5 & 20.8 & 56.3 \\
UniVLA~\cite{univla} & \textbf{83.3} & 66.7 & 33.3 & \textbf{95.8} & 69.8 \\
VITA~\cite{vita} & 68.8 & 95.6 & \underline{84.2} & 37.5 & \underline{71.5} \\
\midrule
\textbf{Ours} & \underline{75.0} & \underline{95.8} & \textbf{87.5} & 33.3 & \textbf{72.9} \\
\bottomrule
\end{tabular}%
}
\label{tab:simpler}
\vspace{-5mm}
\end{table*}

\textbf{SimplerEnv.}
To assess the generalization of our framework, we further evaluate our method on SimplerEnv, with the results summarized in Table~\ref{tab:simpler}.
From the results, our method achieves an average success rate of 72.9\%, surpassing all previous methods. These leading results across diverse benchmarks validate the broad applicability and universality of our framework.

\vspace{-2mm}
\subsection{Ablation Study}
\vspace{-2mm}

Based on the R1–R7 ablations in Table~\ref{tab:ablation_libero}, we observe a consistent trend, showing that each component can improve the performance of our method. Introducing co-training alone demonstrates a substantial gain over the baseline policy (R1→R2/R3, Avg 95.8→97.4/97.9), with co-training with 3D data proving more effective than with 2D data (R3 > R2). To ensure a fair comparison, we incorporate LLAVA-Vision-COCO~\cite{lillava}, a real-world VQA dataset for 2D co-training ablation. This demonstrates that the key limitation of VLA policies lies in insufficient 3D spatial perception and reasoning; thus, enhancing 3D spatial reasoning is beneficial for the action prediction task. Adding the geometry adapter further improves the average success rate (R3→R4, 97.9→98.3), with a notable improvement on LIBERO-Long (93.0→95.8), which suggests that geometry-aware priors are particularly beneficial for long-horizon manipulation tasks with stronger spatial constraints. Subsequently, introducing the reasoning adapter (R4→R5) increases the average score to 98.6, while adding only the reasoning anchor (R4→R6) primarily improves the Goal suite (98.2→99.4), indicating that explicit reasoning anchoring is especially helpful for goal-oriented decision-making. However, its effect is not uniformly distributed across all task categories. The full model (R7: 3D co-training + geometry adapter + reasoning adapter + reasoning anchor) achieves the best overall performance (avg. 98.7) and reaches optimal or joint-optimal results on Spatial/Object/Long, supporting our claim that geometry-aware representation learning and reasoning-aware distillation are complementary. The former provides stable geometrical spatial cues, while the latter strengthens task-relevant reasoning signals for action prediction; their combination leads to the most robust overall behavior. In addition, the margin between R5 and R7 is relatively small (98.6 vs. 98.7), suggesting that the final gain is driven not by a single dominant component, but by cumulative and consistent improvements from multiple modules across different task subsets—consistent with our design principle of training-time enhancement while preserving an efficient inference interface.

\begin{table*}[]
\vspace{-5mm}
\centering
\small
\caption{\textbf{Ablation on Co-training and Adapter Design.} Success rates on LIBERO are presented.}
\vspace{-2mm}
\resizebox{\textwidth}{!}{%
\begin{tabular}{c c c c c | c c c c | c}
\toprule
\textbf{ID} &
\textbf{Cotrain} &
\begin{tabular}[c]{@{}c@{}}\textbf{Geometry}\\\textbf{Adapter}\end{tabular} &
\begin{tabular}[c]{@{}c@{}}\textbf{Reasoning}\\\textbf{Adapter}\end{tabular} &
\begin{tabular}[c]{@{}c@{}}\textbf{Reasoning}\\\textbf{Anchor}\end{tabular} &
\begin{tabular}[c]{@{}c@{}}\textbf{Spatial}\\\textbf{SR(\%)}\end{tabular} &
\begin{tabular}[c]{@{}c@{}}\textbf{Object}\\\textbf{SR(\%)}\end{tabular} &
\begin{tabular}[c]{@{}c@{}}\textbf{Goal}\\\textbf{SR(\%)}\end{tabular} &
\begin{tabular}[c]{@{}c@{}}\textbf{Long}\\\textbf{SR(\%)}\end{tabular} &
\begin{tabular}[c]{@{}c@{}}\textbf{Avg}\\\textbf{SR(\%)}\end{tabular} \\
\midrule
R1 & --      & $\times$ & $\times$ & $\times$ & 93.6 & 99.6 & 97.4 & 92.6 & 95.8 \\
\midrule
R2 & 2D Data & $\times$ & $\times$ & $\times$ & 96.8 & \underline{99.8} & 98.0 & \underline{95.0} & 97.4 \\
R3 & 3D Data & $\times$ & $\times$ & $\times$ & \underline{99.8} & \textbf{100.0} & \underline{98.8} & 93.0 & 97.9 \\
R4 & 3D Data & $\checkmark$ & $\times$ & $\times$ & 99.0 & \textbf{100.0} & 98.2 & \textbf{95.8} & 98.3 \\
R5 & 3D Data & $\checkmark$ & $\checkmark$ & $\times$ & 99.6 & \textbf{100.0} & \underline{98.8} & \textbf{95.8} & \underline{98.6} \\
R6 & 3D Data & $\checkmark$ & $\times$ & $\checkmark$ & 99.2 & \textbf{100.0} & \textbf{99.4} & 94.4 & 98.3 \\
R7 & 3D Data & $\checkmark$ & $\checkmark$ & $\checkmark$ & \textbf{100.0} & \textbf{100.0} & \underline{98.8} & \textbf{95.8} & \textbf{98.7} \\
\bottomrule
\end{tabular}%
}
\label{tab:ablation_libero}
\vspace{-4mm}
\end{table*}

\begin{figure}[t]
    \centering
    \begin{minipage}[b]{0.389\linewidth}
        \centering
        \includegraphics[width=\linewidth]{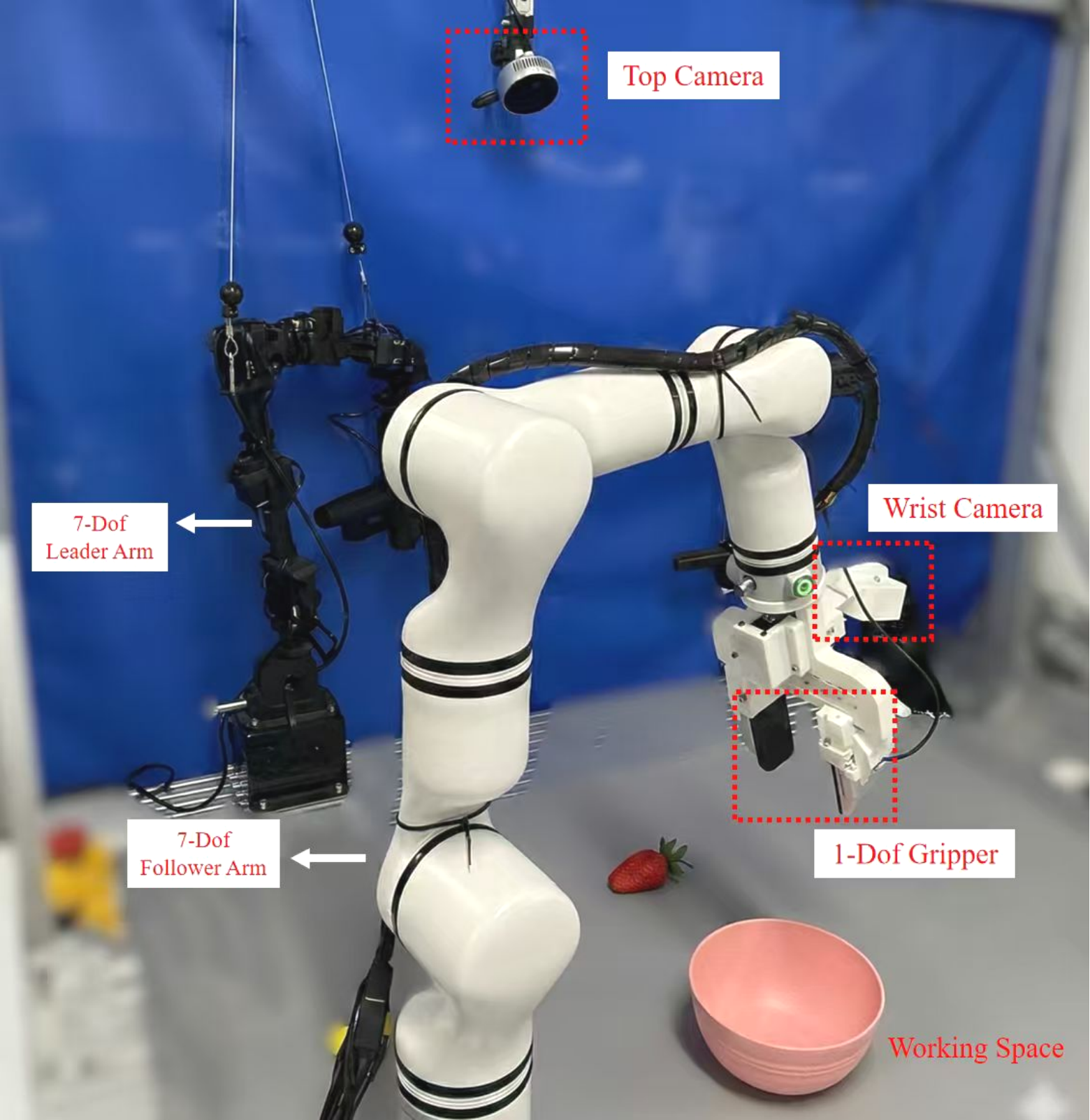}
        \centerline{\small Robot Platform}
    \end{minipage}%
    \begin{minipage}[b]{0.611\linewidth}
        \centering
        \includegraphics[width=\linewidth]{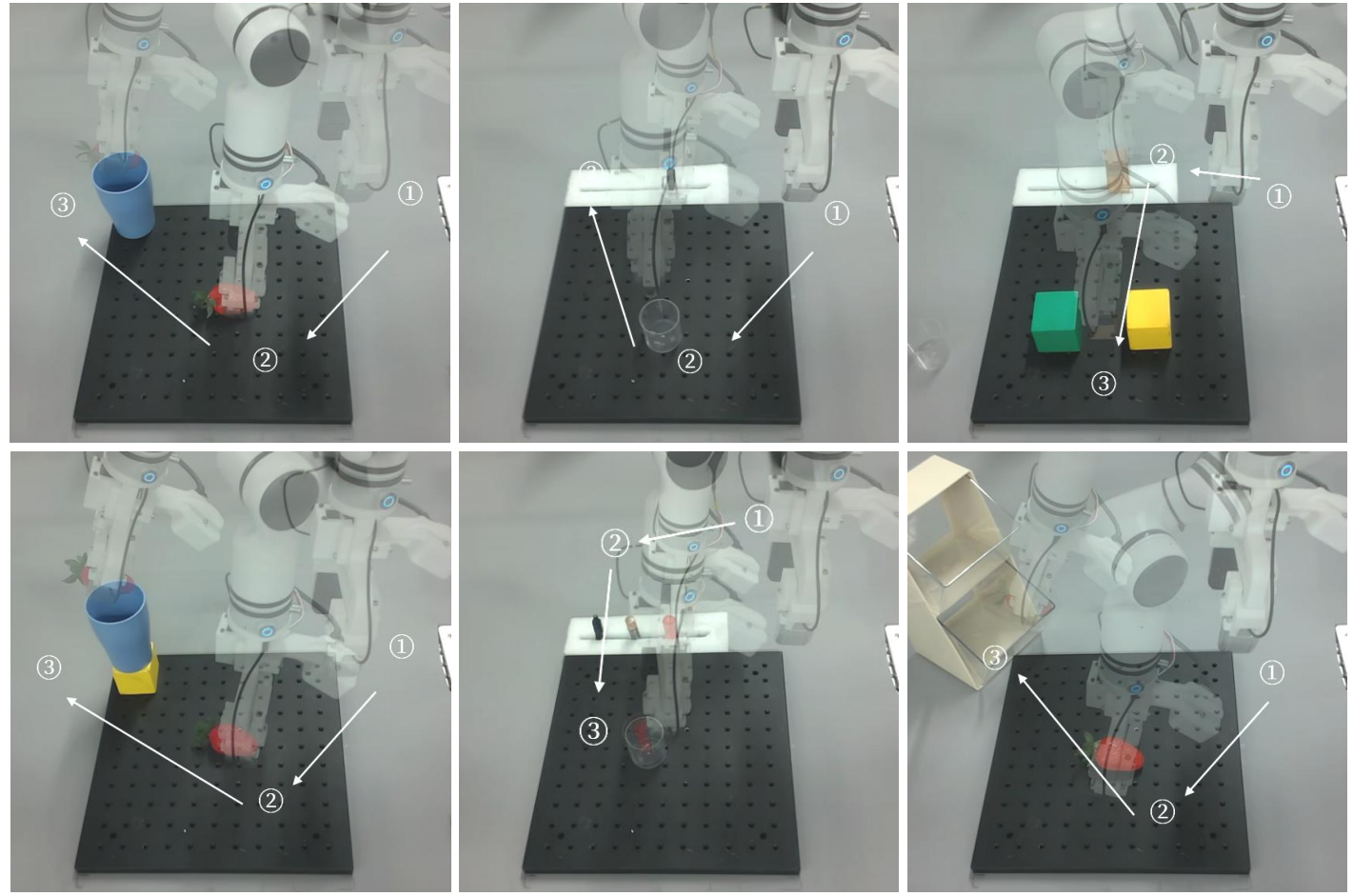}
        \centerline{\small Task 1 \hspace{18mm} Task 2 \hspace{18mm} Task 3}
    \end{minipage}
    \vspace{-5mm}
    \caption{\textbf{Real-world Robot Setup.} (a) Realman robot platform with 7-DoF manipulator, 1-DoF gripper, top camera, and wrist camera. (b) Experimental setups for three real-world tasks: height variation generalization, transparent container placement, and spatial position understanding.}
    \vspace{-6mm}
    \label{fig:robotsetup}
\end{figure}

\begin{wraptable}{r}{0.5\textwidth}
\vspace{-13mm}
\small
\centering
\caption{Success rates on real-world tasks.}
\label{tab:real_results}
\vspace{-1mm}
\begin{tabular}{l c c c }
\toprule
{Methods} & Task 1 & Task 2 & Task 3 \\
\midrule
$\pi_0$~\citep{pi0} & 63.3\% & 82.0\% & 51.3\% \\
OpenVLA-OFT~\citep{openvla-oft} & 28.0\% & 57.3\% & 30.7\% \\
Ours & $\mathbf{88.0\%}$ & $\mathbf{93.3\%}$ & $\mathbf{61.3\%}$ \\
\bottomrule
\end{tabular}
\vspace{-5mm}
\end{wraptable}

\vspace{-2mm}
\subsection{Real-world Evaluation}
\vspace{-2mm}

\textbf{Setup.}
As shown in Figure~\ref{fig:robotsetup}, we perform real-robot experiments on the Realman platform. Each arm is equipped with a 7-DoF manipulator and a 1-DoF gripper, along with a top camera and a wrist camera. We design three tasks to verify the effectiveness of our method: Task 1 evaluates generalization under height variation; Task 2 assesses transparent target recognition by placing objects into transparent containers; Task 3 examines spatial understanding by placing objects at specified spatial positions. All models are trained with 100 episodes per task. For evaluation, we conduct 50 trials per variation for each task (150 trials in total).

\textbf{Results.} 
According to the results displayed in Figure~\ref{fig:robotsetup} and Table~\ref{tab:real_results}, our method demonstrates robust and consistent performance across all scenarios, including height variations, transparent containers, and precise spatial targets. Notably, it significantly outperforms previous methods that lack 3D thinking, validating the necessity of incorporating 3D spatial reasoning for complex manipulation.

\vspace{-3mm}
\section{Conclusion}
\vspace{-3mm}

In this paper, we introduced \OURS, a 3D-thinking-guided co-training framework that enables vision-language-action (VLA) models to perform implicit 3D spatial reasoning using only 2D images. Keys to our method are: (1) disentangling 3D geometry perception and high-level spatial reasoning into hierarchical latent features; (2) an online 3D reasoning distillation module that bridges the prompt-induced reasoning gap via a shared reasoning anchor token, transferring knowledge from explicit reasoning prompts to action-prediction paths; and (3) a spatially augmented action integration that jointly injects these features into the action head as complementary conditions. Extensive experiments on LIBERO, LIBERO-PLUS, SimplerEnv, and real-world benchmarks demonstrate that our approach achieves state-of-the-art performance while maintaining high inference efficiency without the need for 3D sensors or explicit text generation.


{
\small
\bibliographystyle{unsrt}
\bibliography{main}
}


\clearpage
\appendix

\section{Implementation Details}
\label{app:implementation}
We adopt the StarVLA~\cite{ye2026starvla} framework with Qwen3-VL 2B~\cite{bai2025qwen3vltechnicalreport} as the vision-language backbone. We attach an OFT-style action head to predict continuous robot control actions, with all action dimensions normalized during training. Training is conducted on 8 NVIDIA A100 (80GB) GPUs using mixed-precision (bfloat16). We use the AdamW optimizer with parameter-wise learning rates, setting the base learning rate to $2.5 \times 10^{-5}$, a smaller learning rate of $1.0 \times 10^{-5}$ for the Qwen-VL interface layers, and a larger learning rate of $2.0 \times 10^{-4}$ for the action head to enable faster adaptation while stabilizing the pretrained backbone. A cosine learning rate scheduler with a minimum learning rate of $1.0 \times 10^{-6}$ is applied, along with gradient clipping at a maximum norm of $1.0$. The training objective is a weighted combination of VLA loss and VLM loss, where the VLA loss is implemented as L1 distance and the VLM loss is standard cross-entropy, with weights of $1.0$ and $0.1$, respectively. We optionally include alignment and distillation objectives with weights of $0.5$. All evaluations are performed on a single NVIDIA A100 (80GB) GPU.

\section{Simulation Benchmarks}
\label{app:simulation}
We evaluate our method on three widely recognized robotic manipulation benchmarks.
\begin{itemize}

\item\textbf{LIBERO}~\cite{liu2023libero}: This benchmark is designed to evaluate knowledge transfer and policy generalization across four distinct task suites: LIBERO-Spatial, LIBERO-Object, LIBERO-Goal, and LIBERO-Long. We follow the standard protocol for LIBERO, training policies on 50 human-teleoperated demonstrations for each task and evaluating over 50 trials per task (500 trials per suite).
    \item \textbf{LIBERO-PLUS}~\cite{fei2025libero}: This benchmark is designed to evaluate the generalization ability of the model with 10,030 tasks spanning the following: (1) Objects Layout—Confounding objects and target object displacement. (2) Camera Viewpoints - Position, orientation, and field-of-view changes. (3) Robot Initial States - Manipulator initial pose variations; (4) Language Instructions - LLM-based instruction rewriting. (5) Light Conditions—Intensity, direction, color, and shadow variations; (6) Background Textures—Scene and surface appearance changes. (7) Sensor Noise—Photometric distortions and image degradation.
    \item \textbf{SimplerEnv}~\cite{li24simpler}: Following previous methods~\cite{ye2026starvla}, we train the model on SimplerFractal (Google Robot) and Simpler-Bridge (WidowX) datasets and test our model on Simpler-Bridge. For Simpler-Bridge, the evaluation focuses on real-to-sim transfer across tasks.
\end{itemize}

\section{Limitations} 

Due to the integrated co-training and self-distillation mechanisms, our method requires a 50\% increase in computational overhead during training. Specifically, When using 8 A100 GPUs with a batch size of $32$, the training time is $1.5\text{x}$ that of a vanilla VLA model of the same size.

\section{Co-trained 3D Reasoning Data}
\label{sec:cotrain_data}
The 3D reasoning data used for our joint training is derived from the Spatial Reasoner~\cite{ma2025spatialreasonerexplicitgeneralizable3d}. The primary purpose of incorporating this data is to help the model develop stable and accurate explicit 3D perception and spatial reasoning capabilities. This dataset selects unlabeled natural images from the OpenImages dataset and leverages advanced vision foundation models to extract explicit 3D representations of objects. For example, SAM2~\cite{sam2} is used to obtain instance masks, Depth Anything V2~\cite{yang2024depthv2} is used to estimate depth information, and pose estimation algorithms are applied to derive the 3D coordinates and orientation vectors of objects. Subsequently, strict geometric computations are performed based on these coordinates and vectors, such as calculating the 3D Euclidean distance between two objects or computing the angle via cosine similarity of direction vectors. Precise spatial relationship labels, such as “on the left” or "closer," are then derived from these numerical results. To ensure the quality of the training data, a rigorous filtering strategy is applied during dataset construction. This includes removing scenes with excessively dense and overlapping objects, as well as object categories that current vision models cannot reliably segment or estimate poses for. In addition, samples that fall near ambiguous boundary conditions (e.g., cases where a slight angular deviation would lead to entirely different answers) are directly discarded to avoid introducing contradictory supervision signals into the model. Based on these high-quality 3D pseudo-labels, the dataset constructs three progressively more challenging variants: 1) Basic3D-QA, designed purely to train explicit 3D representation extraction, which only queries object positions or performs simple calculations based on known coordinates; 2) SpatialReasoning-QA, a conventional 3D spatial visual question-answering task that directly outputs the final answer.
3) SpatialReasoning-CoT, which requires the model to generate a full chain-of-thought process, including intermediate 3D parsing and geometric computations. In our joint training setup, to prevent catastrophic forgetting in the pretrained general vision-language model, we mix 24,000 carefully selected SR-CoT samples with 24,000 randomly sampled general image-text instruction data from LLaVA during training.

\section{Additional Ablations}
\paragraph{Reasoning Information Fusion.}
\label{sec:reasoing_fusion}
We conduct an ablation study on the fusion mechanisms for integrating reasoning information within the action expert, with results summarized in Table~\ref{tab:fusion}. We investigate three primary fusion strategies: 1) Cross-Attention Fusion: The last hidden state of the reasoning anchor serves as the query ($Q$), while the last hidden states of the action-query tokens function as the key ($K$) and value ($V$).
2) Add Fusion: The last hidden state of the reasoning anchor is directly element-wise added to the last hidden state of the action-query tokens, incorporating a dropout strategy for regularization.
3) Gate Fusion: A gating mechanism is employed to adaptively learn the addition weights, thereby regulating the flow of reasoning information into the action tokens.

\begin{table}[h]
\centering
\caption{Ablation of Reasoning Information Fusion Methods}
\label{tab:fusion}
\begin{tabular}{lccccc}
\toprule
Method & Spatial & Object & Goal & Long & Avg \\ \midrule
Cross-Attn & 99.4 & 99.8 & 99.0 & 93.6 & 98.0 \\
Add & \textbf{100.0} & \textbf{100.0} & 95.8 & \textbf{98.6} & \textbf{98.7} \\
Gate & 99.6 & 99.8 & \textbf{99.2} & 95.0 & 98.4 \\ \bottomrule
\end{tabular}
\end{table}

\paragraph{Distillation.}
To mitigate the impact of the prompt-induced reasoning gap, we explore various distillation strategies. The core objective of distillation is to enable the VLA model to leverage the spatial reasoning capabilities acquired by the VLM through co-training during action prediction.

As shown in Table~\ref{tab:distillation}, we compare the use of the reasoning anchor against privileged information as distillation signals. Inspired by On-Policy Self-Distillation (OPSD)~\cite{zhao2026selfdistilledreasoneronpolicyselfdistillation}, we implement a teacher model that accepts 3D spatial reasoning prompts and generates explicit reasoning text (\textit{e.g.}, 3D object coordinates and relative distances). This text serves as privileged information, prepended to the action prediction prompt. We then apply Kullback-Leibler (KL) divergence to distill the action token distributions between the student model and the teacher model, which utilizes privileged information. 
However, this approach suffers from significant computational overhead due to the explicit generation of reasoning text. Even with a sparse distillation frequency, every 200 training steps, optimization remains challenging. We attribute this to two factors: first, the relatively small size of the teacher model (2B VLM) leads to unstable distillation signals; second, the substantial domain gap between the action space and the text space hinders intuitive alignment. In contrast, our proposed method utilizes the reasoning anchor as an intermediate representation. By treating the anchor as an implicitly privileged token that is shaped by 3D reasoning data, we achieve soft alignment between spatial reasoning and action prediction. This approach is more lightweight, converges faster, and demonstrates superior stability.

\begin{table}[h]
\centering
\caption{Ablation of Distillation Signals}
\label{tab:distillation}
\begin{tabular}{lccccc}
\toprule
Method & Spatial & Object & Goal & Long & Avg \\ \midrule
Baseline & 99.0 & 100.0 & 98.2 & 95.8 & 98.3 \\
Reasoning Anchor & \textbf{100.0} & \textbf{100.0} & 95.8 & \textbf{98.6} & \textbf{98.7} \\
Privileged Info & 99.4 & 99.8 & \textbf{98.6} & 94.4 & 98.1 \\ \bottomrule
\end{tabular}
\end{table}

\paragraph{Explicit vs. Implicit 3D Geometry.}
We further investigate the impact of different 3D geometric information injection methods, as detailed in Table~\ref{tab:3d_info_injection}. Our baseline is the Qwen3-VL-2B model integrated with an OFT-style action head without co-training. For explicit 3D injection, we utilize Point Transformer v3 (PTv3)~\cite{wu2024ptv3} and 3D Diffusion Policy (DP3) encoders~\cite{Ze2024DP3} to extract geometric features from point clouds (predicted via VGGT~\cite{wang2025vggt}), which are then fused with action tokens via addition. To facilitate 3D-free inference and maintain cross-modal balance, we apply a dropout strategy to randomly discard the point cloud modality during training, ensuring the point cloud branch is inactive during inference. Experimental results indicate that injecting explicit 3D geometry is less effective than our implicit reasoning approach. This suggests that the implicit modeling of 3D relationships within the VLM's latent space provides more robust priors for robotic manipulation than raw geometric features.

\begin{table}[h]
\centering
\caption{Comparison of 3D Information Injection Methods}
\label{tab:3d_info_injection}
\begin{tabular}{lccccc}

\toprule
Method & Spatial & Object & Goal & Long & Avg \\ \midrule
Baseline & 93.6 & 99.6 & 97.4 & 92.6 & 95.8 \\
PTv3 & \textbf{98.6} & 99.6 & \textbf{99.2} & 91.6 & 97.3 \\
DP3 Encoder & 93.8 & \textbf{100.0} & 98.0 & 94.2 & 96.5 \\
Geometry Adapter & 98.2 & \textbf{100.0} & 97.4 & \textbf{94.8} & \textbf{97.6} \\ \bottomrule
\end{tabular}
\end{table}

\section{Co-training Gradient Analysis}
\label{sec:gradient}

\begin{figure*}[]
    \centering
    \includegraphics[width=1\linewidth]{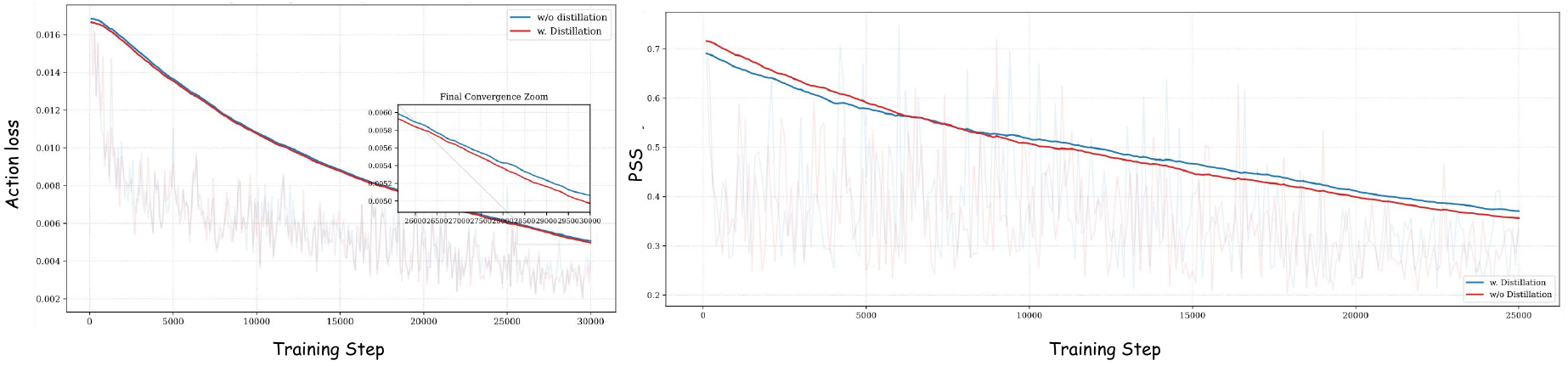}
    \caption{(a) Training action loss curve and (b) Projection-Space Similarity Analysis. {\color{blue}Blue} is utilizing our proposed online 3D reasoning distillation module, and  {\color{red}red} is only 3D co-training.}
    \label{fig:pss}
\end{figure*}

In multi-task co-training, different tasks often share a common backbone network while optimizing distinct objectives. Due to the discrepancy between task-specific optimization targets, the corresponding gradients may interfere with each other in the shared parameter space, leading to gradient conflicts, catastrophic forgetting, and unstable policy learning.
To quantitatively measure the optimization alignment between vision-language modeling and robotic action learning, we adopt Projection-Space Similarity (PSS) as the gradient-level analysis metric.
As shown in Figure~\ref{fig:pss}, co-training with 3D data demonstrates superior stability; the action loss decreases steadily while the PSS remains consistently above zero, maintaining a value of approximately 0.4. This indicates a robust synergistic effect between the tasks. Furthermore, the integration of our proposed Online 3D Reasoning Distillation Module significantly accelerates convergence and enhances the PSS during co-training. These results confirm that our methodology effectively facilitates action prediction and ensures that the optimization gradients remain non-conflicting, thereby successfully mitigating catastrophic forgetting.

Let the shared parameter layer be denoted as $\theta \in \mathbb{R}^{d \times n}$,
where $\theta$ represents the selected bottleneck parameter matrix in the shared VLM backbone. Following prior work, we choose the query projection matrix in the final self-attention layer as the analysis target, since it lies at the intersection of visual-language reasoning and action generation.
For the vision-language modeling task and robotic action task, we define their corresponding losses as
$\mathcal{L}_{\text{vlm}}$, and $\mathcal{L}_{\text{vla}}$. The corresponding gradients with respect to the shared parameters are: 

\begin{equation}
G_{\text{vlm}} = \nabla_{\theta} \mathcal{L}_{\text{vlm}},
\end{equation}

\begin{equation}
G_{\text{vla}} = \nabla_{\theta} \mathcal{L}_{\text{vla}},
\end{equation}
where $G_{\text{vlm}}, G_{\text{vla}} \in \mathbb{R}^{d \times n}$. These gradient matrices characterize the optimization directions induced by the two tasks in the shared parameter space.
Instead of treating gradients as flattened vectors, PSS interprets each gradient matrix as a linear subspace and measures the overlap between the corresponding optimization subspaces.
For a gradient matrix $G$, the associated orthogonal projection matrix is defined as $P = G G^{+}$, where $G^{+}$ denotes the Moore-Penrose pseudoinverse.
Accordingly, the projection matrices for the two tasks are

\begin{equation}
P_{\text{vlm}}
=
G_{\text{vlm}}
G_{\text{vlm}}^{+},
\end{equation}

\begin{equation}
P_{\text{vla}}
=
G_{\text{vla}}
G_{\text{vla}}^{+}.
\end{equation}

\begin{figure*}[]
    \centering
    \includegraphics[width=\linewidth]{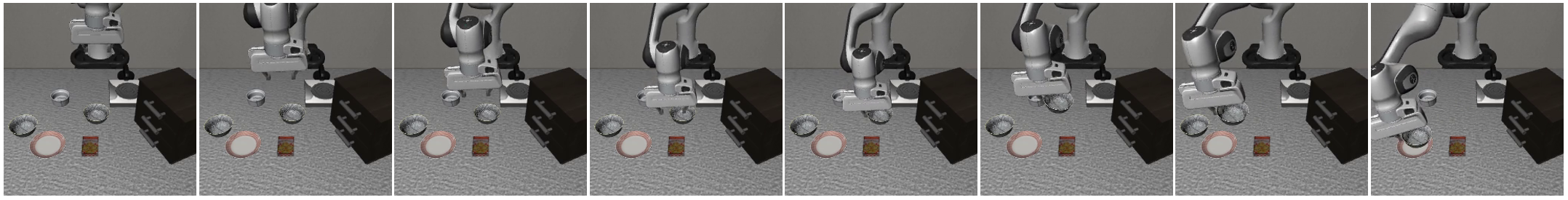}\par\vspace{2mm}
    \includegraphics[width=\linewidth]{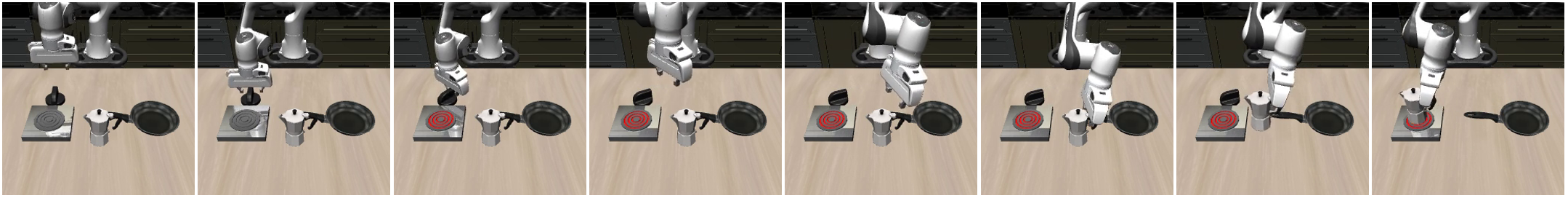}\par\vspace{2mm}
    \includegraphics[width=\linewidth]{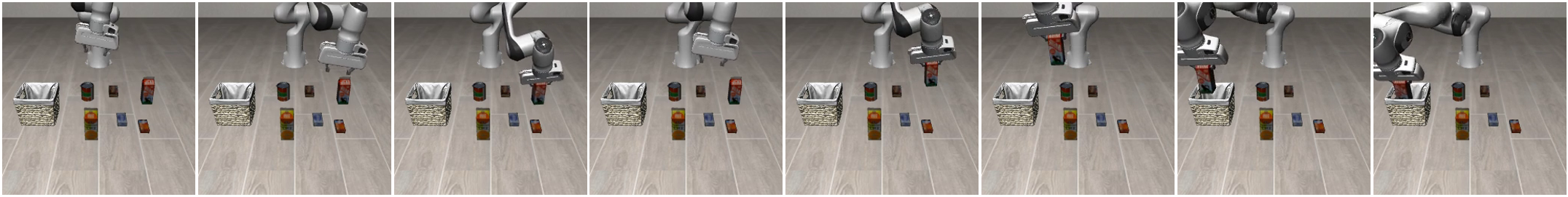}\par\vspace{2mm}
    \includegraphics[width=\linewidth]{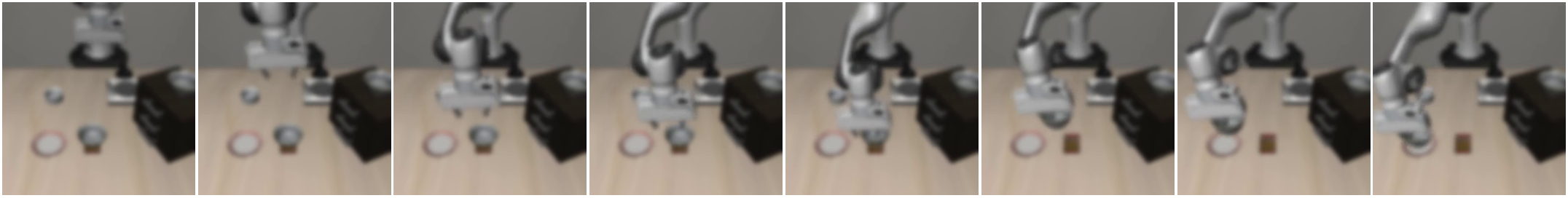}\par\vspace{2mm}
    \includegraphics[width=\linewidth]{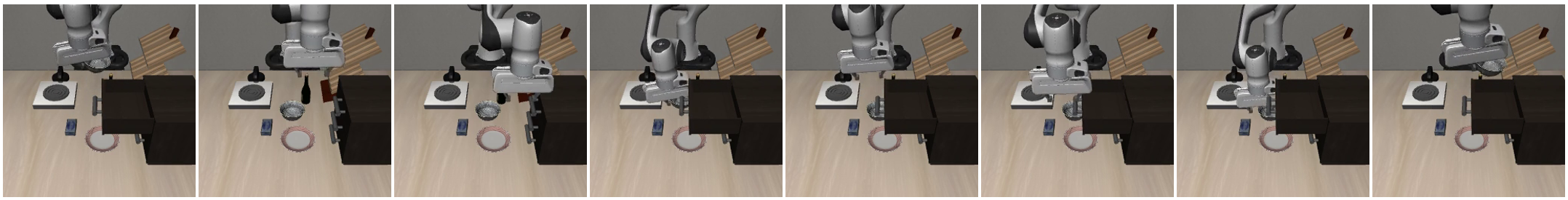}\par\vspace{2mm}
    \includegraphics[width=\linewidth]{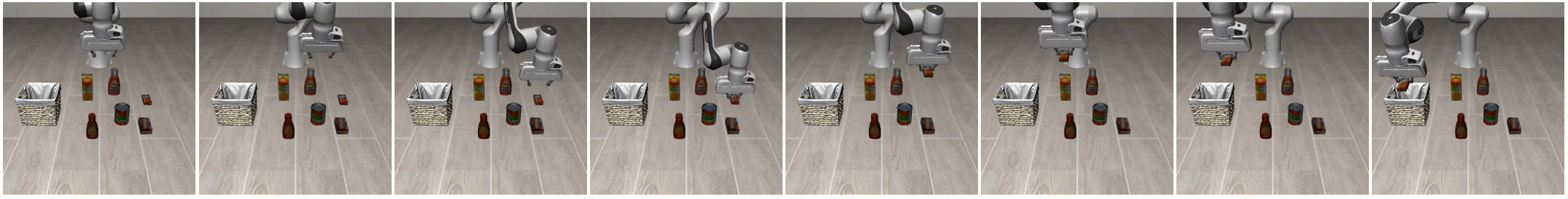}
    \caption{Additional qualitative results on the LIBERO-Plus benchmark.}
    \label{fig:additional_sim1}
\end{figure*}

The Projection-Space Similarity between the two gradient subspaces is then computed as

\begin{equation}
\mathrm{PSS}(G_{\text{vlm}}, G_{\text{vla}})
=
\frac{
\mathrm{Tr}
\left(
P_{\text{vlm}} P_{\text{vla}}
\right)
}{
\min
\left(
\mathrm{rank}(G_{\text{vlm}}),
\mathrm{rank}(G_{\text{vla}})
\right)
}.
\end{equation}

where $\mathrm{Tr}(\cdot)$ denotes the matrix trace operation and $\mathrm{rank}(\cdot)$ denotes matrix rank.
The PSS score satisfies

\begin{equation}
0 \leq \mathrm{PSS} \leq 1.
\end{equation}

A larger PSS value indicates stronger alignment between the optimization subspaces of the two tasks, while lower values imply severe gradient conflicts and parameter competition in the shared backbone.
Compared with conventional gradient similarity metrics, PSS captures the high-dimensional structural consistency between gradient subspaces and therefore provides a more informative characterization of task interactions in large-scale VLA models.

\section{Additional Results}

\begin{figure*}[]
    \centering
    \includegraphics[width=\linewidth]{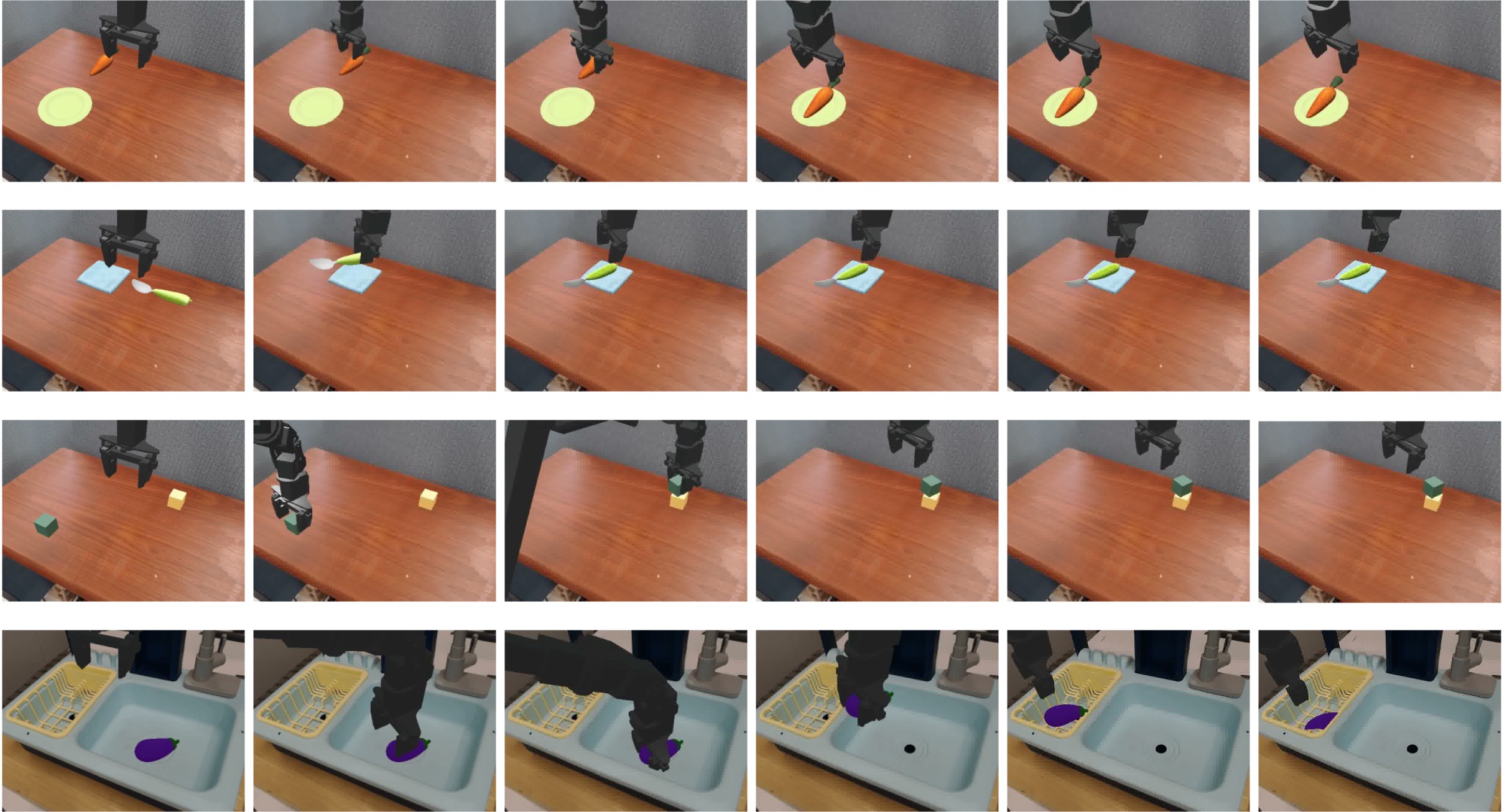}
    \caption{Additional qualitative results on the SimplerEnv benchmark.}
    \label{fig:appendix_simpler}
\end{figure*}

\paragraph{Additional qualitative results.}
To further demonstrate the robustness and versatility of our proposed method, we provide a broader set of qualitative visualizations across various task suites on the LIBERO-Plus benchmark and SimplerEnv benchmark, as displayed in Figure~\ref{fig:additional_sim1} and Figure~\ref{fig:appendix_simpler}.

\begin{figure*}[]
    \centering
    \includegraphics[width=1\linewidth]{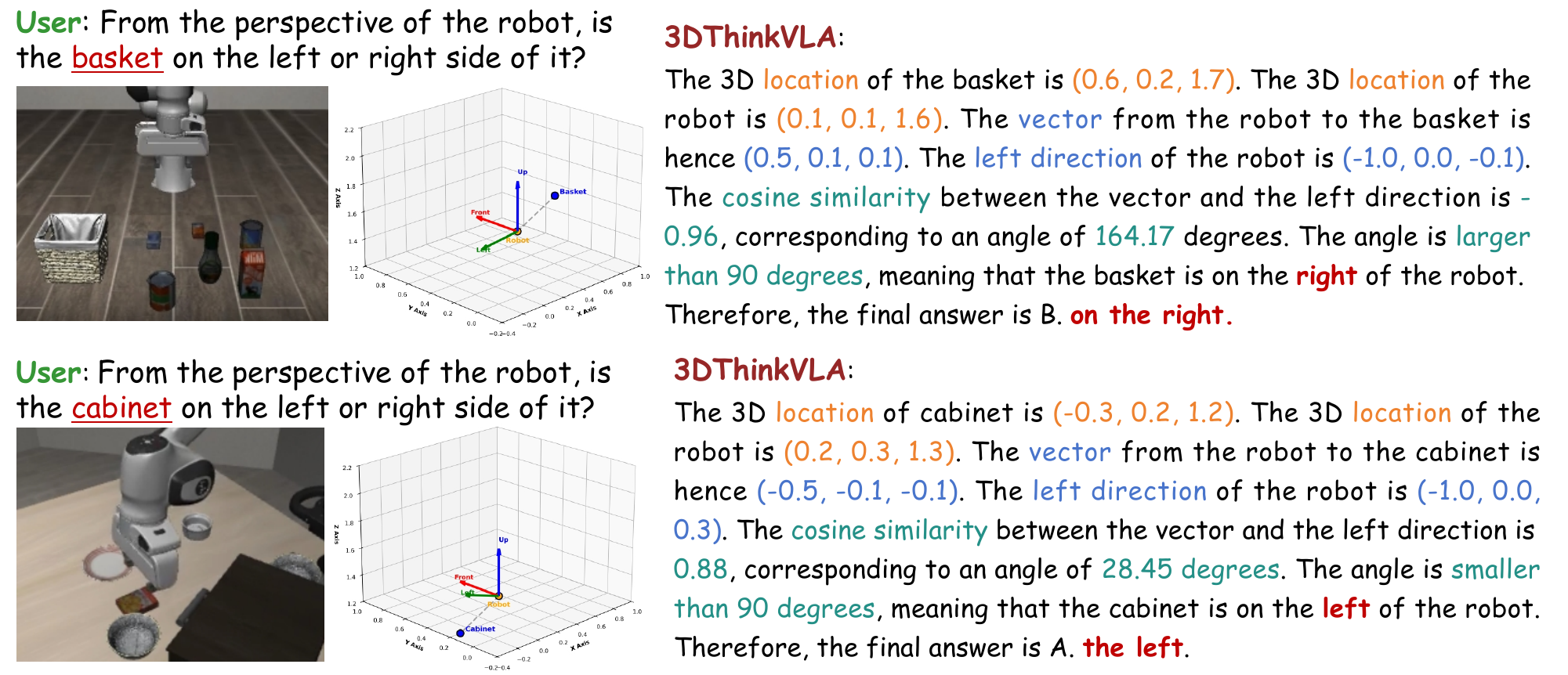}
    \caption{Examples of embodied spatial question-answering tasks on LIBERO (up) and LIBERO-PLUS (down). \OURS can answer embodied spatial reasoning questions through only co-training on real-world 3D data.}
    \label{fig:embodied_qa}
\end{figure*}

\FloatBarrier
\paragraph{Embodied 3D reasoning on the LIBERO and LIBERO-PLUS.}
We display the examples of embodied spatial question-answering tasks on LIBERO and LIBERO-PLUS benchmarks in Figure~\ref{fig:embodied_qa}.
Our method can answer embodied spatial reasoning questions through only co-training on real-world 3D data.

\paragraph{General capabilities of the pretrained VLM.}
Benefiting from our well-designed co-training framework, $\mathcal{L}_{\text{vlm}}$ and $\mathcal{L}_{\text{vla}}$ exhibit a strong synergistic effect during the training phase, as detailed in Appendix~\ref{sec:gradient}. Consequently, the model's generalization capabilities after the training are better preserved. Below, we present several examples of the model performing general-purpose tasks, including code generation, storytelling, and mathematical reasoning:

\definecolor{democolor}{RGB}{160, 44, 44}
\definecolor{usercolor}{RGB}{60, 160, 60}

\newlist{interaction}{description}{1}
\setlist[interaction]{
    labelsep=0.5em,
    leftmargin=1.5em,
    itemsep=1.5ex,
    topsep=1ex,
    style=nextline,
    font=\small\bfseries\sffamily
}

\begin{interaction}

    \item[{\color{usercolor}User}:] 
        Write an example C++ code.

    \item[{\color{democolor}3DThinkVLA}:] 
        Sure! Here's an example C++ program that prompts the user to enter their name and then prints a greeting:

        \begin{small}
        \begin{verbatim}
#include <iostream>
#include <string>

int main() {
    std::cout << "Enter your name: ";
    std::string name;
    std::getline(std::cin, name);
    std::cout << "Hello, " << name << "!" << std::endl;
    return 0;
}
        \end{verbatim}
        \end{small}

    \item[{\color{usercolor}User}:] 
        Tell me a story.

    \item[{\color{democolor}3DThinkVLA}:] 
        Once upon a time, in a far-off kingdom, there was a beautiful princess named Rose... 
        [\dots] 
        The kingdom mourned the loss of their beloved princess and prince, but they knew that Rose and Leo's love would live on in the memories of all who knew them. And so, their story became a legend, a reminder to all of the power of true love and the happiness it can bring.

    \item[{\color{usercolor}User}:] 
        If the line $y = 2x + 5$ is tangent to the curve $y = e^x + x + a$, then $a = ?$

    \item[{\color{democolor}3DThinkVLA}:] 
        If the line is tangent to the curve, it means that there is exactly one point where the line intersects the curve. Setting the two equations equal to each other:
        \[ 2x + 5 = e^x + x + a \]
        Simplifying, we get:
        \[ e^x - x = 5 - a \]
        Taking the derivative of both sides with respect to $x$ to find the point of tangency:
        \[ e^x - 1 = 0 \implies x = 0 \]
        Substituting $x = 0$ back into the equality $2x + 5 = e^x + x + a$:
        \[ 2(0) + 5 = e^0 + 0 + a \implies 5 = 1 + a \]
        Therefore, $a = 4$.

\end{interaction}



\end{document}